\documentclass[]{article}
\usepackage{amssymb}
\usepackage{amsmath}
\usepackage{amsthm}
\title{ Compact Shape Trees:  A Contribution to the Forest of Shape Correspondences and Matching Methods }
\author{Abdulrahman Oladipupo Ibraheem \\ rahmanoladi@yahoo.com \\ \\ Computing and Intelligent Systems Research Group \\ Department of Computer Science and Engineering \\ Obafemi Awolowo University, Ile-Ife, Nigeria. }
\begin{document}
\maketitle
\begin{abstract}
We propose a novel technique, termed \textbf{compact shape trees}, for computing correspondences of single-boundary 2-D shapes in $O(n^2)$ time. Together with zero or more features defined at each of $n$ sample points on the shape's boundary, the compact shape tree of a shape comprises the $O(n)$ collection of vectors emanating from \textbf{any} of the sample points on the shape's boundary to the rest of the sample points on the boundary. As it turns out, compact shape trees have a number of elegant properties both in the spatial and frequency domains. In particular, via a simple vector-algebraic argument, we show that the $O(n)$ collection of vectors in a compact shape tree possesses at least the same discriminatory power as the $O(n^2)$ collection of lines emanating from \textbf{each} sample point to every other sample point on a shape's boundary. In addition, we describe neat approaches for achieving scale and rotation invariance with compact shape trees in the spatial domain; by viewing compact shape trees as aperiodic discrete signals, we also prove scale and rotation invariance properties for them in the Fourier domain. Towards these, along the way, using concepts from differential geometry and the Calculus, we propose a novel theory for sampling 2-D shape boundaries in a scale and rotation invariant manner. Finally, we propose a number of shape recognition experiments to test the efficacy of our concept. 
\end{abstract}
\section{Introduction}
Shape recognition is a consequential field of computer vision which studies, and seeks to automate, the processes by which previously known shapes can be ``known again." Indeed, its importance is well underscored by how dated its mention occurs in the literature. For instance, Smith(2001) transmits the following implicit reference to it from \textit{Kitab al-Mannazir}, an eleventh century book written by Ibn Al-Haytham, a prominent scholar of the Islamic Golden Age:
\begin{quotation}
\noindent \textit{Sight also perceives many things by means of recognition, so it recognizes that a human is a human, that a horse is a horse, ... when it has seen the same thing before.}
\end{quotation} 
Till date, research into shape recognition continues to receive significant attention from computer scientists, cognitive psychologists, neuroscientists, as well as mathematicians. One famous method that was proposed quite recently \textemdash at the turn of the last decade \textemdash is shape contexts (Belongie, 2002). Shape contexts use a two-dimensional histogram of point counts to establish the correspondence between two shapes, and to compute a similarity score between the shapes. This concept paper proposes a shape descriptor that is very similar in spirit to shape contexts. As a tool for shape recognition, we propose the \textbf{compact shape tree}. Together with zero or more features defined at each of $n$ sample points on the shape's boundary, the compact shape tree of a shape comprises the $O(n)$ collection of vectors emanating from \textbf{any} of the sample points on the shape's boundary to the rest of the sample points on the boundary. Indeed, Belongie \textit{et. al.} already contemplated the use of the collection of the \textbf{lines} ( not vectors) emanating from any of the sample points on the shape's boundary to the rest of the sample points on the boundary as an means for computing shape correspondences, towards recognition. However, they jettisoned it on grounds that it is likely to be unstable to intra-class variations. Albeit, the recent work of Wang \textit{et. al.}, who employed a very close variant of compact shape trees with success on well known shape databases, indicates that compact shape trees should be viable for shape recognition.
Yet, there is a major difference between the approach taken here and those of Belongie \textit{et. al.}, and Wang \textit{et. al.}. To compute the correspondences between two shapes, the approaches of those authors must "walk through" every sample point on the second shape for each sample point on the first, thereby resulting in an $O(n^3)$ time algorithm, where $n$ is the number of sample points on each shape. To the contrary, our method walks through every sample point on the second shape for \textbf{only one} \textemdash indeed \textbf{any one} \textemdash sample point on the first. Our algorithm begins by choosing any compact shape tree on the first shape. It then walks through all the sample points on the second shape, comparing the compact shape tree rooted at each point with the chosen compact shape tree of the first shape. The best matching compact shape tree on the second shape is then used to establish correspondences between the sample points of the two shapes. This results in an $O(n^2)$ time algorithm. 
It turns out that compact shape trees have a number of quite interesting properties. First, although the vectors in a compact shape tree form an $O(n)$ collection, they possess at least the same discriminatory power as \textbf{shape forests}, our terminology for the $O(n^2)$ collection of lines emanating from \textbf{each} sample point to every other sample point on a shape's boundary. This is because the length and angles of each line on the shape forest can be computed from the vectors comprising the compact shape tree. In fact, it should be possible to show that the compact shape tree is both a minimal and minimum "subset" of a shape forest with this property.
More so, compact shape trees yield naturally to simple translation, scale, and rotation invariance techniques both in the frequency and spatial domains. In both domains, translation invariance is an intrinsic property of compact shape trees. In the frequency domain, we view the moduli and angles associated with the vectors comprising a compact shape tree as constituting aperiodic discrete signals. We then show that the scaled Fourier transforms of these signals is invariant to rotation and scaling. In the spatial domain, following Al-Ajlan, we achieve scale invariance simply by scaling the modulus of each vector in the compact shape tree by the maximum of the moduli. To achieve rotation invariance, we take the difference between the angles of consecutive \textemdash traversing the underlying shape's boundary in a counter-clockwise manner \textemdash vectors in the compact shape tree. . 
\section{Compact Shape Trees for Computing Correspondences}
We start by looking at compact shape trees consisting of vectors alone. Figures 2a and 2b depict two shapes, each with eight sample points. Respectively, the sample points are ${p_1, p_2,...,p_8}$ and ${q_1, q_2,...,q_8}$; in practice, hundred sample points are typically used for better accuracy, so that the scenarios in Figures 2a and 2b are merely illustrative. Undoubtedly, the two shapes in the figures look very similar. If compact shape trees are anything to count on, then they must be able to capture the similarity between the two shapes. In Figure 2c, we exhibit, for the shape in Figure 2a, a compact shape tree rooted at point $p_8$. On the other hand, in Figure 2d, we depict for the shape in Figure 2b a compact shape tree rooted at point $q_8$. It is obvious that the compact shape trees in Figures 2c and 2d look very similar; they capture the similarity between the two underlying shapes. Indeed, if one were to consider any other compact shape tree on the shape in Figure 2b, one would find that none matches the compact shape tree of Figure 2c the way the compact shape tree of Figure 2d does. For instance, if we consider the compact shape tree rooted at $q_4$, shown in Figure 2e, we see clearly that it is poles apart from the structure in Figure 2c. Thus, it is only intuitive to conclude that the best matching compact shape tree, obtainable from the shape of Figure 2b, for the compact shape tree of Figure 2c is that in Figure 2d. From these pair of compact shape trees, one can easily obtain the correspondences: $p_1 \leftrightarrow q_1, p_2 \leftrightarrow q_2, ... p_8 \leftrightarrow q_8$, where ''$\leftrightarrow$" stands for "corresponds to." We obtain $p_8 \leftrightarrow q_8$ from the root of the trees, and obtain the remaining correspondences from the "tips" of the vectors. 
Let us now see what happens, if we consider any other compact shape tree on Figure 2a. Suppose we consider the compact shape tree rooted at $p_4$ (shown in Figure 2e). We should find that the best match to the tree of Figure 2e, from the compact shape trees of Figure 2b, is the one rooted at $q_4$. We show this compact shape tree in Figure 2f. Now, when we try to retrieve the correspondences between the compact shape trees of Figures 2e and 2f, we get: $p_1 \leftrightarrow q_1, p_2 \leftrightarrow q_2, ... p_8 \leftrightarrow q_8$, which is exactly what we obtained when we used the tree rooted at $p_8$ on the first tree. Based on this observation, we intuit that the choice of compact shape tree on the first shape should be immaterial; this is the most salient contribution of this paper, and this is what allows our method achieve its $O(n^2)$ run time property.
Following Kimia et.al., Belongie et. al. and Wang et. al., we are going to parlay tangency/curvature information into our model. We believe that tangency/curvature is a key visual feature that should be taken into account by any scheme aimed at the computation of correspondences. In what follows, we describe, at a rather formal level, how to match compact shape trees consisting of the vectors we have seen so far, along with curvature vectors and their moduli, measured at boundary sample points. 
Let $\mathcal{C_P}$ and $\mathcal{C_Q}$ be twice differentiable simple closed curves defining the boundaries of two arbitrary shapes in the plane. In a counter-clockwise fashion, we mark equally-spaced sample points on each of the curves, obtaining for $\mathcal{C_P}$, the set $\mathcal{P} = {p_1, p_2,...,p_n}$, and for $\mathcal{C_Q}$ the set $\mathcal{Q} = {q_1, q_2,...,q_n}$, where $n$ is typically taken as $100$. Next, we introduce $\mathcal{L}_{p_i} = {p_i, p^1, p^2, ...,p^{n-1}}$, which we call a \textbf{rooted labeling} of $\mathcal{P}$ \textbf{with respect to} $p_i$. $\mathcal{L}_{p_i}$ defines the order in which sample points are encountered on $\mathcal{C_P}$, when traversed in a counter-clockwise manner, starting at $p_i$. We denote the compact shape tree rooted at $p_i$ by $\mathcal{T}_{p_i}$. Evidently,$\mathcal{T}_{p_i}$ comprises $n-1$ vectors. We label these vectors, $u_1, u_2,...,u_{n-1}$, such that: the "tip" of $u_i$ is the sample point $p^i$. Analogously, we may speak of tree $\mathcal{T}_{q_i}$ with $n-1$ vectors: $v_1, v_2,...,v_{n-1}$. Further, for each $i \in {1, \dots, n-1}$, the curvature at: $p^i$ is denoted $\delta_i$; the curvature at $p_i$, the root of $\mathcal{T}_{p_i}$, is denoted $\delta_n$. Similarly, the curvature at each $q^i$ is denoted $\kappa_i$, while that at $q_i$ is denoted $\kappa_n$. With these, we can now define a \textbf{tentative} cost, $\hat{C}(\mathcal{T}_{p_i}, \mathcal{T}_{q_i})$, of matching $\mathcal{T}_{p_i}$ and $\mathcal{T}_{q_i}$ as: 
\begin{equation}
\hat{C}(\mathcal{T}_{p_i}, \mathcal{T}_{q_i}) = w_1 \sum_{i=1} ^{n-1} |u_i -v_i|^2 + w_2 \sum_{i=1} ^{n} (\delta_i -\kappa_i)^2
\end{equation} 
In the above equation, $w_1$ and $w_2$ are weights, which we leave up to training data to set for us. The first term of the equation measures a weighted sum of the square magnitudes of the differences between corresponding vectors in $\mathcal{T}_{p_i}$ and $\mathcal{T}_{q_i}$, while the second term measures a weighted sum of square Euclidean distances between curvatures. In preliminary experiments, we have noticed a problem with the discriminatory power of the second term of the above equation; we describe one way of fixing the problem in the next section.
For now, we describe how to use the above equation. Of all compact shape trees from $\mathcal{Q}$, the one that best matches the tree rooted at $p_i$, is denoted $\mathcal{T}_{\hat{q}}$, where $\hat{q}$ is the minimizer of $\hat{C}(\mathcal{T}_{p_i}, \mathcal{T}_{q_i})$:
\begin{equation}
\hat{q} = \underset{q_i \in \mathcal{Q}}{\textup{argmin}} \; \hat{C}(\mathcal{T}_{p_i}, \mathcal{T}_{q_i})
\end{equation} 
To retrieve the point correspondences being sought, we denote by $\hat{q}, \hat{q}^1, \hat{q}^2 ...,\hat{q}^{n-1}$ the rooted labeling of $\mathcal{Q}$ with respect to $\hat{q}$. We imagine simultaneously traversing curves $\mathcal{C_P}$ and $\mathcal{C_Q}$, starting at $p_i$ and $\hat{q}$ respectively. The other in which we encounter sample points on both curves establishes the correspondences being sought. As depicted in Figure 2, we find: $\hat{q}\leftrightarrow p_i$ at the roots of $\mathcal{T}_{\hat{q}}$ and $\mathcal{T}_{p_i}$; and $\hat{q}^1\leftrightarrow p^1$, $\hat{q}^2\leftrightarrow p^2$, ...,$\hat{q}^{n-1}\leftrightarrow p^{n-1}$ at the tips of the vectors in $\mathcal{T}_{\hat{q}}$ and $\mathcal{T}_{p_i}$. 
\section{A Problem With Distance Sums in Euclidean Space: the Curvature Case}
Herein, we describe the results of preliminary experiments that we performed on the curvature functions of ellipses. Consider Figure ELLIP. There are two half-ellipses in the figure, and the two of them can be parametrized with respect to angle $\theta$: $x = a \cos \theta$ and $y = b \sin \theta$. The first half-ellipse has parameters $a = 3$ and $b = 7$, while the second ellipse has $a = 7$ and $b = 3$. We now ask: which half-ellipse is most similar in ``shape" to the straight line shown in the figure? Many human observers will definitely not say that both ellipses have the same level of similarity to the straight line. To our utmost dismay however, under zero-space sampling for continuous curves, the second term of Equation 1 says they are. Although, we don not normally use zero-space sampling in practice, and more so the curves we work with in practice are not continuous, towards drawing lessons from it, it is nonetheless worthwhile to examine this problem in some depth. When we looked closely at the problem, we found it is does not stem from the choice of curvature as a feature used in Equation 1. Rather, it is a consequence of the phenomenon of distinct yet equidistant points in Euclidean space. Put clearly, the reason why the second term of Equation 1 says that both half-ellipses have the same level of similarity to the straight line is that both half-ellipses are equidistant to the straight line in Euclidean curvature space. It turns out this issue is more mundane than it might at first seem, because it can be traced to the commutativity and associativity of the elementary addition operation: 2 + 3 + 10 = 10 + 2 + 3.
It is appropriate that we describe our findings at a technical level. To begin with, in Mokhtarian and Mackworth (1992), the curvature function $\kappa(t)$ of a curve, $\mathcal{C}(t) = (x(t), \: y(t))$, parametrized by variable $t$, is shown to be given by:
\begin{equation} 
\kappa(t) = \dfrac{x_t(t)y_{tt}(t)- y_t(t) x_{tt}(t)}{(x_t(t)^2 + y_t(t)^2)^\frac{3}{2}}
\end{equation}
As earlier hinted at, we parametrized the half-ellipses by angle $\theta$: $x(\theta) = a \cos \theta$ and $y(\theta) = b\sin \theta$, so that $x_{\theta}(\theta) = -a \sin \theta$, $x_{\theta \theta}(\theta) = -a \cos \theta$, $y_{\theta}(\theta) = b \cos \theta$, and $y_{\theta \theta}(\theta) = -b \sin \theta$. Combining these, and allowing $\theta$ to play the role of $t$ in Equation 3 above, we get:
\begin{equation} 
\kappa(\theta) = \dfrac{ab \sin^2 \theta \: + \: ab \cos^2 \theta }{(a^2 \sin^2 \theta \: + \: b^2 \cos^2 \theta)^\frac{3}{2}}
\end{equation}
Using the identity, $\sin^2 \theta \: + \: \cos^2 \theta = 1$, we find:
\begin{equation} 
\kappa(\theta) = \dfrac{ab }{(a^2 \sin^2 \theta \: + \: b^2 \cos^2 \theta)^\frac{3}{2}}
\end{equation}
Advancing, we denote the curvature function of the straight line shown in Figure ELLIP by $\kappa_0$, and $\kappa_0$ is everywhere zero, the line being straight. Further, we denote the curvature functions of the first and second half-ellipses by $\kappa_1(\theta)$ and $\kappa_2(\theta)$ respectively. Keeping in mind that $a = 3 $ and $b = 7$ for the first half-ellipse, then according to the second term of Equation 1, the ``difference" between the first half-ellipse and the straight line is given by: 
\begin{equation} 
D_1 = w_2 \int_{\frac{-\pi} {2}} ^{\frac{\pi} {2}} (\dfrac{21}{(3^2 \sin^2 \theta \: + \: 7^2 \cos^2 \theta)^\frac{3}{2}} \: - \: 0)^2 \, d \theta
\end{equation}
In the same vein, the difference, as measured by the second term of Equation 1, between the second ellipse and the straight line is:
\begin{equation} 
D_2 = w_2 \int_{\frac{-\pi} {2}} ^{\frac{\pi} {2}} (\dfrac{21}{(7^2 \sin^2 \theta \: + \: 3^2 \cos^2 \theta)^\frac{3}{2}} \: - \: 0)^2 \, d \theta
\end{equation}
When we performed the integrations numerically using MATLAB function \textit{integral}, we found $D_1 = D_2$, which implies $w_2D_1 = w_2 D_2$. In fact, we found that this is always the case for any pair of half-ellipses, such that the first one has parameters $a = a_1, b = b_1$, and the second one has parameters $a = b_1, b = a_1$. In Figure ELLIPPLOTa, we plot a graph of $D_2 - D_1 + 5$ against $\theta$ for the half-ellipses of Figure ELLIP. As can be seen, the graph is everywhere equal to $5$, an indication that $D_1 = D_2$, and by extension $w_2D_1 = w_2 D_2$. In Figure ELLIPPLOTb, we plot a similar graph for another pair of half-ellipses, the first having $a = 17, b = 69$, and the second having $a = 69, b = 17$. Again, we find that the graph stays at $5$ through out.
The key issue is how to deal with the problem described above. Our investigations revealed that one possible solution is to employ the natural logarithm of the first order moments of the squared distances between the curvature functions. We denote this $M$, and express it as:
\begin{equation}
M = \log _e (\int_{t_1} ^{t_2} t \, [\: \kappa_p(t)-\kappa_q(t) \:]^2 \, dt)
\end{equation} 
where $\kappa_p(t)$ is the curvature function on the boundary of the first shape and $\kappa_q(t)$ is the curvature function on the boundary of the second shape.
We applied the above Equation to compare the two half-ellipses of Figure ELLIP with the straight line, and found that it is able to ``discriminate" between them. In particular, for the first ellipse we find $M =$, while for the second ellipse we found $M =$. Furthermore, we observed that the quantity $M$ increases as the ``protrusion'' of the half-ellipses increase. For example, $M$ is greater for the second half-ellipse of Figure ELLIP than it is for the first, because the second half-ellipse is more protruded than the first. We present data to corroborate this observation in Table 1.
The above observations suggest that we replace the matching cost, $\hat{C}(\mathcal{T}_{p_i}, \mathcal{T}_{q_i})$, defined in Equation 1, with a new cost, $C(\mathcal{T}_{p_i}, \mathcal{T}_{q_i})$, defined as follows: 
\begin{equation}
C(\mathcal{T}_{p_i}, \mathcal{T}_{q_i}) = w_1 \sum_{i=1} ^{n-1} |u_i -v_i|^2 + w_2 \sum_{i=1} ^{n} (\delta_i -\kappa_i)^2 + \log_e( w_3 \sum_{i=1} ^{n} i (\delta_i -\kappa_i)^2) 
\end{equation} 
The first two terms of the above equation are exactly the same as the two terms of Equation 1. The third term of the equation can be considered a discrete version of the right hand side of Equation 8. Notice that, in the third term, the index $i$ plays the role parameter $t$ plays in the continuous version given in Equation 8; it is acceptable to parametrize discrete curves by indices. 
\section{One Theory and One Conjecture about Compact Shape Trees}
In the previous section, we stated that when comparing two shapes, only one compact shape tree on the first shape needs to compared with the compact shape trees of the second shape. Is there any formal theoretical guarantee that there is no match leading to a different set of correspondences amongst the $n-1$ ``uncompared" compact shape trees on the first shape? As suggested by Figure 1, we believe there is none. However, because we do not yet have a proof at this time, we state this observation as a formal conjecture:
\newtheorem{conj1}{Conjecture}
\begin{conj1}
Let $\mathcal{P} $and $\mathcal{Q}$ respectively denote sets of $n$ sample points on two shapes in the plane. For a given compact shape tree, $\mathcal{T}_{p_i} = \{p_i,p^1,p^2,...,p^{n-1}\} \subseteq \mathcal{P}$, suppose, amongst all the compact shape trees obtainable from $\mathcal{Q}$, the one that best matches $\mathcal{T}_{p_i}$ is the tree $\mathcal{T}_{\hat{q}} = \{\hat{q},q^1,q^2,...,q^{n-1}\} \subseteq \mathcal{Q}$. Further suppose, for some other compact shape tree, $\mathcal{T}_{r} = \{r,r^1,r^2,...,r^{n-1}\} \subseteq \mathcal{P}, r \neq p_i$, amongst all the compact shape trees obtainable from $\mathcal{Q}$, the one that best matches $\mathcal{T}_{r}$ is the tree $\mathcal{T}_{\hat{s}} = \{\hat{s},s^1,s^2,...,s^{n-1}\} \subseteq \mathcal{Q}$. Then, the correspondences, $p_i \leftrightarrow \hat{q}, p^1 \leftrightarrow q^1, p^2 \leftrightarrow q^2, ..., p^{n-1} \leftrightarrow q^{n-1}$, retrieved from matching $\mathcal{T}_{p_i}$ and $\mathcal{T}_{\hat{q}}$ are exactly the same as (i.e. just a permutation of) the correspondences, $r \leftrightarrow \hat{s}, r^1 \leftrightarrow s^1, r^2 \leftrightarrow s^2, ..., r^{n-1} \leftrightarrow s^{n-1}$, retrieved from matching $\mathcal{T}_{r}$ and $\mathcal{T}_{\hat{s}}$. 
\end{conj1} 
Also, in the Introduction section, we asserted that the vectors in a compact shape tree have at least the same discriminatory power as a shape forest, which is the $O(n^2)$ collection of lines emanating from each sample point on a shape's boundary to every other sample point on the boundary. We are now going to prove this claim after casting it formally thus:
\newtheorem{thm1}{Theorem}
\begin{thm1}
Respectively, let $\mathcal{T}_{\hat{p}}$ and $\mathcal{F}$ be a compact shape tree and a shape forest obtained from $n$ sample points of a simple closed curve. Further, let $l_k$ and $\beta_k$ be the length and angle, with respect to the positive $x$-axis, of the $k$-th line in $\mathcal{F}$. Each $(l_k, \beta_k)$ pair can be obtained from the vectors in $\mathcal{T}_{\hat{p}}$, either directly or via calculation, so that all the information in the $O(n^2)$ collection, $\mathcal{F}$, is also available in the $O(n)$ collection of vectors in $\mathcal{T}_{\hat{p}}$. 
\end{thm1}
\begin{proof}[\textbf{\textit{Proof}}] \mbox{}\\
Consider any two distinct points $p_i$ and $p_j$ in $\mathcal{F}$, and denote by $l_k$ and $\beta_k$ the length and angle, with respect to the positive $x$-axis, of the line joining $p_i$ and $p_j$ in $\mathcal{F}$. There are only three possibilities: \textup{1)}. $p_i = \hat{p}$, so that $p_j \neq \hat{p}$ \quad \textup{2)}. $p_j = \hat{p}$, so that $p_i \neq \hat{p}$ \quad \textup{3)}. $p_i \neq \hat{p}$, and $p_j \neq \hat{p}$ as well. For the first possibility, the directed edge $(p_i, p_j)$ can be written as the directed edge $(\hat{p}, p_j)$, which obviously is a vector, $v$, in $\mathcal{T}_{\hat{p}}$. Thus, $l_k$ and $\beta_k$ can be obtained from the polar representation of $v$. The second possibility is analogous to the first one. For the third possibility, we note that $p_i$ and $p_j$ must be the tips of two distinct vectors, $v_1$ and $v_2$, in $\mathcal{T}_{\hat{p}}$. Now, we simply set $v_3 = v_2 - v_1$, which says that $v_3$ is a vector extending from $p_i$ to $p_j$. It is clear that $l_k$ and $\beta_k$ can then be obtained from the polar representation of $v_3$. 
\end{proof}
\section{Theories and Algorithms for Scale and Rotation Invariant Sampling }
In the previous sections, we spoke severally of sampling $n$ points of the boundary of a shape. What exactly is the best way of choosing these sample points ? Although, Ling et. al., Wang et. al., and Belongie et. al have used boundary sample points in their works, they do not explicitly describe any algorithms for computing them, except that they said that the sample points are equally spaced along the boundary of the shape. In particular, they do not mention whether or not their sample points have been computed in a scale and rotation invariant manner, whereas it would be optimal to employ scale and rotation invariant sample points for computing shape correspondences. In this section, we are going to describe three possible approaches to computing sample points. The first approach is an heuristic which comes quite close to being rotation and scale invariant, while the latter two are based on theories from basic differential geometry and Calculus. We shall prove rigorously that, under fair assumptions, these latter two techniques lead to exact scale and rotation invariant sampling. \\
We want to define what we mean by scale and rotation invariant sampling. In $\mathbb{R}^2$, we allow the operator $\mathcal{E}_{\gamma}(.)$ to denote the operation of enlarging a shape by a factor of $\gamma$. Also, we use $R_{\phi}$ to denote the $2$ by $2$ matrix that abstracts rotation of shapes by $\phi$ degrees in the plane, and use $R_{\phi}(.)$ to denote this operation. Hence, if shape $\tilde{\mathcal{S}}$ is the result of first enlarging another shape, $\mathcal{S}$, by a factor of $\gamma$, and then rotating this intermediate result by $\phi$, one may write: $\tilde{\mathcal{S}} = R_{\phi} (\: \mathcal{E}_{\gamma}(\mathcal{S}) \: )$. Now, where $\mathcal{P} = {p_1, p_2, \dots , p_n}$ is a sampling of $n$ points on the boundary of ${\mathcal{S}}$, and $\tilde{\mathcal{P}} = {\tilde{p}_1, \tilde{p}_2, \dots , \tilde{p}_n}$ is a sampling of $n$ points on the boundary of $\tilde{\mathcal{S}}$, then we say that the samplings are rotation and scale invariant if ${\mathcal{P}}$ and $\tilde{\mathcal{P}}$ satisfy the bijection $f: {\mathcal{P}} \rightarrow \tilde{\mathcal{P}}$, in which $f(.) = R_{\phi} (\: \mathcal{E}_{\gamma}(.) \: )$. Put simply, if the samplings are rotation and scale in invariant, then each point $p_i \in \mathcal{P}$ is mapped by $R_{\phi} (\: \mathcal{E}_{\gamma}(.) \: )$ to a point $\tilde{p}_j \in \tilde{\mathcal{P}}$, and the point $\tilde{p}_j$ is mapped back to the point $p_i$ by $R_{\phi}^{-1} (\: \mathcal{E}_{ \frac{1}{\gamma} }(.) \: )$, $R_{\phi}^{-1}$ being the inverse of $R_{\phi}$. \\ 
From the above definition, it would appear that, after sampling $\mathcal{S}$ to obtain $\mathcal{P}$, then one can obtain $\tilde{\mathcal{P}}$ simply by applying the operation $R_{\phi} (\: \mathcal{E}_{\gamma}(.) \: )$ on the elements of $\mathcal{P}$. However, this can lead to a vicious cycle: many approaches obtain $R_{\phi} (\: \mathcal{E}_{\gamma}(.) \: )$ by solving the shape correspondence problem between ${\mathcal{S}}$ and $\tilde{\mathcal{S}}$, but to solve this shape correspondence problem, we need both $\mathcal{P}$, and $\tilde{\mathcal{P}}$ ready. As we shall see, our theory of invariant sampling leads to an algorithm that does not need any knowledge of $R_{\phi} (\: \mathcal{E}_{\gamma}(.) \: )$ to work. 
\subsection{A Pseudo-Invariant Heuristic for Shape Boundary Sampling} 
We describe a pseudo-invariant sampling algorithm that works on an heuristic rule. The algorithm expects as inputs a closed curve $\mathcal{C}$, which forms the boundary of some shape $\mathcal{S}$, as well as the number, $n$, of sample points to be computed along $\mathcal{C}$. The condition $n = 2^k$, for some non-negative integer $k$, must be met, because the "hidden" objective of the algorithm is to recursively divide $\mathcal{C}$ into two \textbf{equal} halves until a stopping criterion is met. Although, ideally, the algorithm should output equally-spaced sample points, it turns out this is not the case for digital curves, due to the effects of digitization. To illustrate the essence of the method, we take shape $\mathcal{S}$ to be a disk having a circumference of $24$cm. Also, we take $\gamma = 2$ and $\phi = 90^{\circ}$ (actually, we know rotation does not have any real effect on circles), so that $\tilde{\mathcal{S}} = R_{90} (\: \mathcal{E}_{2}(\mathcal{S}) \: )$ is simply the disk whose circumference is $48$cm. When $n = 8$, for disk $\mathcal{S}$, the algorithm computes a set of eight sample points $\mathcal{P} = {p_1, p_2, \dots , p_8}$, such that the arc length, $s$, between each pair of consecutive sample points on the disk's boundary is given by $s = \dfrac{24}{8} = 3$cm. Likewise, with $n = 8$, for the second disk, $\tilde{\mathcal{S}}$, the algorithm obtains a set of eight sample points, $\tilde{\mathcal{P}} = {\tilde{p}_1, \tilde{p}_2, \dots , \tilde{p}_8}$, such that the arc length, $s$, between each pair of consecutive sample points on the disk's boundary is given by $s = \dfrac{48}{8} = 6$cm. In Figure 2, we show both disks and the sample points computed for them by the algorithm. From the figure, it is easy to see that a bijection, $f(.) = R_{\frac{\pi}{2}} (\: \mathcal{E}_{2}(.) \: )$, readily exists between ${\mathcal{P}}$ and $\tilde{\mathcal{P}}$. Notice that this says that the algorithm has been able to compute rotation and scale invariant samplings for the disks. More importantly, it should not be hard to see that the algorithm will always compute rotation and scale invariant for any non-digitized shape. \\
Albeit, we have identified two drawbacks of the algorithm. First, the requirement $n = 2^k$ is a very harsh one. For instance, suppose we wish to sample $100$ points, then, we must settle for either $2^6 = 64$ points, or $2^7 = 128$ points. Clearly, neither $64$ nor $128$ is a good approximation to $100$. Thus, the condition $n = 2^k$ embodies a really stringent constraint. The second drawback concerns digitalized shapes. This problem stems from the fact that pixel counts are always integers, so when the algorithm needs to pick the middle point of an open digital curve having a length of $24$ pixels it must either pick the twelfth or thirteenth pixel. But, definitely, both pixels are neither the exact middle of the open curve. Well, we believe this does not lead to terrible deviations from the invariant behavior of the algorithm we saw in the case of continuous curves. Thus, the must significant drawback is the first one. The next two sub-sections are devoted to discussions of theories and algorithms which do not suffer this drawback. 
\subsection{A Theory of Invariant Boundary Sampling by Distance Minimization: Part I}
The style of this sub-section will be largely formal. But, before the formalisms, we will give an overview of the technique being proposed. At base, the proposed technique harnesses the observation that distances from a shape's boundary points to the shape's centroid are invariant to rotation, and are simply multiplied by the scaling factor, when the shape is scaled and/or rotated. That is, if $D(s)$ is function that measures these distances before scaling and/or rotation, and $\tilde{D}(\tilde{s})$ is function that measures them after scaling and/or rotation, then the functions satisfy the relationship $\tilde{D}(\tilde{s}) = \gamma D(s)$, where $\gamma$ is the scaling factor. Using techniques from Calculus, we shall study some theoretical and practical implications of the above relationship. Specifically, we will show that the minima/maxima of $D(s)$ \textbf{correspond} to the minima/maxima of $\tilde{D}(\tilde{s})$, under the rule $\tilde{s} = \gamma s$. Practically, this means, for example, that if $s_1$ is the only absolute minimum point of $D(s)$ then $\tilde{s}_1 = \gamma s_1$ will be the only absolute minimum of $\tilde{D}(\tilde{s})$. In this scenario, we can harness $s_1$ and $\tilde{s}_1$ as \textbf{seed points} in our boundary sampling algorithm; the rest of the sample points are calculated with respect to $s_1$ and $\tilde{s}_1$ respectively. Furthermore, as we shall see, by solving a simple distance-correspondence problem around the boundaries of the shapes, we are able to extend our algorithm to cover the case wherein $D(s)$ and $\tilde{D}(\tilde{s})$ each have more than one absolute minima/maxima. We can now begin the formalism: 
\newtheorem{thm9}{Definition} 
\begin{thm9}
Let $\mathcal{S}$ be any shape in $\mathbb{R}^2$, and let $\tilde{\mathcal{S}}$ be the shape that results when $\mathcal{S}$ is enlarged by a factor of $\gamma$ and then rotated through $\phi$ radians in the plane. We shall express this as $\tilde{\mathcal{S}} = \mathcal{R}_{\phi} (\: \mathcal{E}_{\gamma}(\mathcal{S}) \: )$, where $\mathcal{E}_{\gamma}(.)$ denotes the enlargement operation, $\mathcal{R}_{\phi}(.)$, denotes the rotation operation, and $R_{\phi}$ is the orthogonal rotation matrix associated with $\phi$. Also, for any $ p \in \mathcal{S}$, and any $\tilde{p} \in \tilde{\mathcal{S}}$, if $\mathcal{R}_{\phi} (\: \mathcal{E}_{\gamma}(.) \: )$ sends $p$ to $\tilde{p}$, then we shall write $\tilde{p} = \mathcal{R}_{\phi} (\: \mathcal{E}_{\gamma}(p) \: )$. Finally, if the curves ${\mathcal{C}}$, and $\tilde{\mathcal{C}}$ are respectively the boundaries of ${\mathcal{S}}$ and $\tilde{\mathcal{S}}$, then we shall indicate this via the expression $\tilde{\mathcal{C}} = \mathcal{R}_{\phi} (\: \mathcal{E}_{\gamma}({\mathcal{C}}) \: )$
\end{thm9}
\newtheorem{thm2}[thm1]{Theorem} 
\begin{thm2}
Suppose $\mathcal{C}$ and $\tilde{\mathcal{C}}$ are two twice-differentiable simple closed curves in $\mathbb{R}^2$, such that $\tilde{\mathcal{C}} = R_{\phi} (\: \mathcal{E}_{\gamma}(\mathcal{C}) \: )$. Further, on $\mathcal{C}$, let $s_1$ be the arc length, measured counter-clockwise from some arbitrary point $p_i$ to another arbitrary point $p_j$; and, on $\tilde{\mathcal{C}}$, let $\tilde{s}_1$ be the arc length measured counter-clockwise from some arbitrary point $\tilde{p}_i$ to another arbitrary point $\tilde{p}_j$. The following hold: \textup{1).} If $\tilde{p}_i = R_{\phi} (\: \mathcal{E}_{\gamma}(p_i) \: )$ \textbf{and} $\tilde{p}_j = R_{\phi} (\: \mathcal{E}_{\gamma}(p_j) \: )$, then we must have $ \tilde{s}_1 = \gamma s_1 $. \textup{2).} if $\tilde{p}_i = R_{\phi} (\: \mathcal{E}_{\gamma}(p_i) \: )$ \textbf{and} $ \tilde{s}_1 = \gamma s_1 $, then we must have $\tilde{p}_j = R_{\phi} (\: \mathcal{E}_{\gamma}(p_j) \: )$. \\ \\
\end{thm2}
\begin{proof}[\textbf{\textit{Proof}}] \mbox{}\\
In what follows, on $\mathcal{C}$, arc length, $s$, is measured counter-clockwise from $p_i$; while, on $\tilde{\mathcal{C}}$, arc length, $\tilde{s}$, is measured counter-clockwise from $\tilde{p}_i$. 
Beginning with the first part of the theorem, assume $\tilde{p}_i = R_{\phi} (\: \mathcal{E}_{\gamma}(p_i) \:)$ and $\tilde{p}_j = R_{\phi} (\: \mathcal{E}_{\gamma}(p_j) \:))$. Parameterizing by $s$, one may write $\mathcal{C}(s) = ( \: x(s), \; y(s) \: )^T$ (Here, we are deliberately using \textbf{column} vectors to make certain matrices compatible for multiplication). Further, with $(\: \tilde{x}(s), \; \tilde{y}(s) \:)^T$ representing the point which $R_{\phi} (\: \mathcal{E}_{\gamma}(.)$ sends $( \: x(s), \; y(s) \: )^T$ to, we have $( \: \tilde{x}(s), \; \tilde{y}(s) \: )^T = R_{\phi} \gamma ( \: x(s), \; y(s) \: )^T$, and $\tilde{\mathcal{C}}(s) = ( \: \tilde{x}(s), \; \tilde{y}(s) \: )^T = R_{\phi} \gamma ( x(s), \; y(s))^T$.
Now, the quantity $s_1$ can be obtained from, $s_1 = \int_0 ^{s_1} ds$. To obtain $\tilde{s}_1$, we start with the relationship, $d\tilde{s} = |\tilde{T}(s)| ds$, in which ${T}(s)$ denotes the tangent vector, as a function of $s$, on curve $\tilde{\mathcal{C}}(s)$. One can compute $\tilde{T}(s)$ by differentiating $\tilde{\mathcal{C}}(s)$ with respect to $s$: $\tilde{T}(s) = \tilde{\mathcal{C}}_s(s) = R_{\phi} \gamma ( \: x_s(s), \; y_s(s) \: )^T$. To compute the magnitude of $\tilde{T}(s)$, we first write the rotation matrix, $R_{\phi}$, explicitly as: $R_{\phi} = \begin{pmatrix}\cos \phi & \sin \phi \\ - \sin \phi & \cos \phi \end{pmatrix} $. This allows us to see: $\tilde{T}(s) = \begin{pmatrix}\cos \theta & \sin \theta \\ - \sin \theta & \cos \theta \end{pmatrix} \gamma \begin{pmatrix} x_s(s) \\ y_s(s) \end{pmatrix} =$ $\begin{pmatrix} \gamma x_s \cos \phi \: + \: \gamma y_s \sin \phi \\ - \gamma x_s \sin \phi \: + \: \gamma y_s \cos \phi \end{pmatrix}$, from which we obtain, $|\tilde{T}(s)| = ( \: ( \gamma x_s \cos \phi \: + \: \gamma y_s \sin \phi)^2 \: + \: ( - \gamma x_s \sin \phi \: + \: \gamma y_s \cos \phi)^2 \: )^{\dfrac{1}{2}} \: = \: (\: \gamma^2 x_s^2 \cos^2 \phi + 2\gamma^2 x_s y_s\cos \phi \sin \phi + \gamma^2 y_s^2 \sin^2 \phi + \gamma^2 x_s^2 \sin^2 \phi - 2\gamma^2 x_s y_s\cos \phi \sin \phi + \gamma^2 y_s^2 \cos^2 \phi \: )^{\dfrac{1}{2}}$. Using the identity, $\sin^2 \phi + \cos^2 \phi = 1$, we ultimately find: $|\tilde{T}(s)| = \gamma ( \: x_s(s)^2 \: + \: y_s(s)^2 \:)^{\dfrac{1}{2}}$. But, the quantity $( \: x_s(s)^2 \: + \: y_s(s)^2 \:)^{\dfrac{1}{2}}$ equals $|\mathcal{C}_s(s)|$. Moreover, since $s$ is the arc length parameter of $\mathcal{C}(s)$, we must have $|\mathcal{C}_s(s)| = 1$. In symbols we write, $|\tilde{T}(s)| = \gamma |\mathcal{C}_s(s)| = \gamma$. Now, if one plugs this into the relationship, $d\tilde{s} = |\tilde{T}(s)| ds$, one gets: $d\tilde{s} = \gamma ds$. Finally, integrating this from $s = 0$ to $s = s_1$, we get $\tilde{s}_1 = \int_0 ^{s_1} \gamma ds = \gamma s_1$.
For the second part of the theorem, we prove by contradiction. Assume $\tilde{p}_i = R_{\phi} (\: \mathcal{E}_{\gamma}(p_i) \: )$ \textbf{and} $ \tilde{s}_1 = \gamma s_1 $, but $\tilde{p}_j \neq R_{\phi} (\: \mathcal{E}_{\gamma}(p_j) \: )$. But, $R_{\phi} (\: \mathcal{E}_{\gamma}(.)$ must map $p_j$ to some point, say $\tilde{p}_k$, on the curve $\tilde{\mathcal{C}}$. Hence, there exists a point $\tilde{p}_k \neq p_j$, such that $\tilde{p}_k = R_{\phi} (\: \mathcal{E}_{\gamma}(p_j) \: )$. By the first part of this theorem, it means that the arc length of $\tilde{p}_k$ is $\tilde{s}_1$. But, this is a contradiction because the two distinct points $\tilde{p}_k$ and $\tilde{p}_j$ cannot have the same arc length. 
\end{proof}
The above theorem paves the way to the following theoretical result: 
\newtheorem{thm4}[thm1]{Theorem} 
\begin{thm4}
In $\mathbb{R}^2$, suppose $\mathcal{C}$ and $\tilde{\mathcal{C}}$ are twice differentiable simple closed curves, such that $\tilde{\mathcal{C}} = R_{\phi} (\: \mathcal{E}_{\gamma}(\mathcal{C}) \: )$; and suppose point $p_i \in \mathcal{C}$ and point $\tilde{p}_i \in \tilde{\mathcal{C}}$ are related by $\tilde{p}_i = R_{\phi} (\: \mathcal{E}_{\gamma}(p_i) \: )$. Further, for any two points $p_j \in \mathcal{C}$ and $\tilde{p}_j \in \tilde{\mathcal{C}}$ related by $\tilde{p}_j = R_{\phi} (\: \mathcal{E}_{\gamma}(p_j) \: )$, let $s$ be the arc length of $\mathcal{C}$, measured (counter) clockwise from $p_i$, and let $\tilde{s}$ be the arc length of $\tilde{\mathcal{C}}$, measured (counter) clockwise from $\tilde{p}_i$. Then, where $s_t$ is the total arc length of $\mathcal{C}$, and where $\tilde{s}_t$ is the total arc length of $\tilde{\mathcal{C}}$, if $G: [0, s_t]/ \{0,s_t\} \rightarrow \mathbb{R} $ and $\tilde{G}: [0, \tilde{s}_t]/ \{0,\tilde{s}_t\} \rightarrow \mathbb{R} $ are two functions satisfying $\tilde{G}(\tilde{s})= cG(s)$ for some positive constant, $c$, the following must hold. \textup{1).} $s_1$ is an absolute minimum point of $G(s)$ if and only if $R_{\phi} (\: \mathcal{E}_{\gamma}(s_1))$ is an absolute minimum point of $\tilde{G}(\tilde{s})$. 
\textup{2).} $s_1$ is an absolute maximum point of $G(s)$ if and only if $R_{\phi} (\: \mathcal{E}_{\gamma}(s_1))$ is an absolute maximum point of $\tilde{G}(\tilde{s})$.
\end{thm4}
\begin{proof}[\textbf{\textit{Proof}}] \mbox{} \\
Theorem 2 enforces $\tilde{s} = \gamma s$. We can use this to map sub-intervals of $I = [0, s_t]/ \{0,s_t\}$ to sub-intervals of $\tilde{I} = [0, \tilde{s}_t]/ \{0,\tilde{s}_t\}$, and vice versa. For example, $[s_0, s_2] \subset I$ is mapped to $[\tilde{s}_0, \tilde{s}_2] \subset \tilde{I}$ under the condition that $\tilde{s}_0 = \gamma s_0$ and $\tilde{s}_2 = \gamma s_2$. With this in mind, we proceed. We recall from Calculus the definition of an absolute minimum of function $G(s)$ on interval $I$. If $s_1$ is an absolute minimum of $G(s)$ on $I$, then $G(s_1) \leq G(s)$ for all $s \in I$. Since $c > 0$, we have $cG(s_1) \leq cG(s) \Rightarrow \tilde{G}(\tilde{s}_1) \leq \tilde{G}(\tilde{s})$ for all $s \in I$. As discussed above, the interval $I$ maps to the interval $\tilde{I}$, since $0 = \gamma 0$ and $\tilde{s}_t = \gamma s_t$. Consequently, we may declare: $\tilde{G}(\tilde{s}_1) \leq \tilde{G}(\tilde{s})$ for all $s \in I $, which says that $\tilde{s}_1$ is an absolute minimum of $\tilde{G}(\tilde{s})$ on $\tilde{I}$. This proves the forward implication of the "if and only if" phrase in the first conclusion of the theorem. For sheer completeness, we spell out a proof of the reverse implication. Assume $\tilde{s}_1$ to be an absolute minimum of $\tilde{G}(\tilde{s})$ on $\tilde{I}$, so that $\tilde{G}(\tilde{s}_1) \leq \tilde{G}(\tilde{s}) \Rightarrow \dfrac{1}{c} \tilde{G}(\tilde{s}_1) \leq \dfrac{1}{c} \tilde{G}(\tilde{s}) \Rightarrow G(s_1) \leq G(s)$ on $\tilde{I} \Rightarrow G(s_1) \leq G(s)$ on $I$, the last implication being a consequence of mapping $\tilde{I}$ to $I$. The proof of the second conclusion of the theorem is analogous to the theorem's first conclusion, so do not write it out. 
\end{proof}
With the aid of the just proven theorem, we can now describe a preliminary version of our scale and rotation invariant sampling algorithm. As we have been doing, we suppose $\mathcal{S}$ and $\tilde{\mathcal{S}}$ are two shapes satisfying $\tilde{\mathcal{S}} = \mathcal{R}_{\phi} (\: \mathcal{E}_{\gamma}(\mathcal{S}) \: )$, and that $\mathcal{C}$ and $\tilde{\mathcal{C}}$ are the boundaries of $\mathcal{S}$ and $\tilde{\mathcal{S}}$ respectively. Further, on $\mathcal{C}$, an arbitrary point $p_i$ is chosen from which arc lengths, are measured, say, counter-clockwise. Similarly, on $\tilde{\mathcal{C}}$, the point $\tilde{p}_i$ satisfying $\tilde{p}_i = \mathcal{R}_{\phi} (\: \mathcal{E}_{\gamma}(\mathcal{p}_i) \: )$ is chosen for the same purpose. Arc lengths are denoted $s$ on $\mathcal{C}$, and denoted $\tilde{s}$ on $\tilde{\mathcal{C}}$. We now let $D(s)$ denote the distance from the centroid of $\mathcal{S}$ to the point with arc length $s$ on $\mathcal{C}$. In the same vein, we let $D(s)$ denote the distance from the centroid of $\tilde{\mathcal{S}}$ to the point with arc length $\tilde{s}$ on $\tilde{\mathcal{C}}$. As mentioned at the opening of this sub-section $D(s)$ and $\tilde{D}(\tilde{s})$ are related by $\tilde{D}(\tilde{s}) = \gamma D(s)$. Therefore, by Theorem 4, the absolute minima/maxima of $D(s)$ must correspond to the absolute minima/maxima of $\tilde{D}(\tilde{s})$, under the rule $\tilde{s} = \gamma s$. The ``preliminary version" of our algorithm is for the case wherein both $D(s)$ and $\tilde{D}(\tilde{s})$ each have a either a single absolute minimum point or a single absolute maximum point. The algorithm outputs a set $\mathcal{P} ={p_1, p_2, .., p_n}$ of $n$ sample points when invoked with $\mathcal{S}$ and $n$; and outputs a set $\tilde{\mathcal{P}} = {\tilde{p}_1, \tilde{p}_2, \dots, \tilde{p}_n}$ of $n$ when called with $\tilde{\mathcal{S}}$ and $n$. 
Now, suppose $s_1$ and $\tilde{_s}_1$ are respectively the ``single points" at which $D(s)$ and $\tilde{D}(\tilde{s})$ reach their absolute minima. Further, suppose ${\mathcal{C}}$ and $\tilde{\mathcal{C}}$ have total arc lengths $s_t$ and $\tilde{s}_t$ respectively. Our algorithm simply takes $s_1$ and $\tilde{_s}_1$ as \textbf{seed points}, and this means $s_1$ and $\tilde{s}_1$ correspond to the first sample points computed for ${\mathcal{C}}$ and $\tilde{\mathcal{C}}$. Next, the algorithm computes the quantities $w = \dfrac{s_t}{n}$ and $\tilde{w} = \dfrac{\tilde{s}_t}{n}$. With these, for $\mathcal{C}$, it is able to compute $\mathcal{P} ={p_1, p_2, .., p_n}$, such that $p_1 \in \mathbb{R}^2$ corresponds to the seed point on $\mathcal{C}$, and each $p_i \in \mathbb{R}^2$, $i \in {2, \dots, n}$, is at an arc length distance of $w$ from $p_{i-1}$. Clearly, this leads to equal-spaced sampling of curve $\mathcal{C}$; See Figure. Analogously, for $\tilde{\mathcal{C}}$, the algorithm is able to compute $\tilde{\mathcal{P}} = {\tilde{p}_1, \tilde{p}_2, \dots, \tilde{p}_n}$ such that $\tilde{p}_1 \in \mathbb{R}^2$ corresponds to the seed point on $\tilde{\mathcal{C}}$, and each $\tilde{p}_i \in \mathbb{R}^2$, $i \in {2, \dots, n}$, is the point that is at an arc length distance of $w$ from $\tilde{p}_1$. Similar to the case of $\mathcal{C}$, this leads to equal-spaced sampling of curve $\tilde{\mathcal{C}}$. 
We look at the situation in some more details. First, Theorem 4, says that $p_1$ and $\tilde{p}_1$ must satisfy $\tilde{\mathcal{\tilde{p}_1}} = \mathcal{R}_{\phi} (\: \mathcal{E}_{\gamma}(\mathcal{p_1}) \: )$. Further, Theorem 2 implies $\tilde{s}_t = \gamma s_t$, which further implies $\tilde{w} = \dfrac{\tilde{s}} {n} = \gamma \dfrac{{s}} {n} = \gamma w$, so that each $\tilde{p}_i$ is at an arc length distance of $\gamma (i-1){w}$ from $\tilde{p}_1$, while, as earlier noted, each $p_i$, $i \in {2, \dots, n}$, is at an arc length distance of $(i-1)w$ from $p_1$. This, together with the fact $\tilde{\mathcal{\tilde{p}_1}} = \mathcal{R}_{\phi} (\: \mathcal{E}_{\gamma}(\mathcal{p_1}) \: )$ as well as the second part of Theorem 2, now inform us that for each $i \in {2, \dots, n}$, the relationship $\tilde{\mathcal{\tilde{p}_i}} = \mathcal{R}_{\phi} (\: \mathcal{E}_{\gamma}(\mathcal{p_i}) \: )$ is in force. Consequently, our ''preliminary algorithm," achieves rotation and scale invariant sampling. Furthermore, it also obliterates the harsh $n = 2^k$ requirement of the method described in the previous sub-section. On the flip side though, in its present form, the algorithm is able to deal only with continuous curves. However, it should not be very difficult to adapt it for digital curves. In particular, this will involve the use of techniques specialized for the estimation of arc length on digital curves. Fortunately, there exist several such techniques in the literature (). We outline our preliminary algorithm as Algorithm 1 below. 
\subsection{A Theory of Invariant Boundary Sampling by Distance Minimization: Part II}
In this sub-section, we attempt to extend our curve boundary sampling algorithm to cover cases wherein the distance functions, $D(s)$ and $\tilde{D}(\tilde{s})$, do not have unique points of absolute minimum/maximum. Our approach will involve solving a simple arc length correspondence problem around the boundaries of the involved curves at recognition time. To this end, let $A = \{a_1, a_2, \dots, a_L\}$, $L > 1$, $a_i \in mathbb{R}^2$, be the set of points at which $D(s)$ reaches absolute minimum, and let $\tilde{A} = \{\tilde{a}_1, \tilde{a}_2, \dots, \tilde{a}_L\}$ be the set of points at which $\tilde{D}(s)$ reaches absolute minimum. An implication of Theorem 4 is that there exists a bijection, $f$ from set $A$ to set $\tilde{A}$. Our goal is to find this bijection. To illustrate this goal, consider Figure X, a depiction of two shapes $\mathcal{S}$ and $\tilde{\mathcal{S}}$, satisfying $\tilde{\mathcal{C}} = R_{\phi} (\: \mathcal{E}_{\gamma}(\mathcal{C}) \: )$. A glance at the figure reveals that the bijection, $f$, we seek can be written in an ordered set $ \mathcal{F}$ as: $ \mathcal{F} = \{a_1 \leftrightarrow \tilde{a}_3, a_2 \leftrightarrow \tilde{a}_1, a_3 \leftrightarrow \tilde{a}_2 \}$, where the ``order" in $\mathcal{F}$ comes from the fact that the correspondences in it are written in counter clockwise fashion around the boundaries of $\mathcal{S}$ and $\tilde{\mathcal{S}}$. By Theorem 2, there is a natural relationship between the arc lengths of \textbf{consecutive} members of $ \mathcal{F}$. For instance, if we consider the second and third members of $\mathcal{F}$, that is, $a_2 \leftrightarrow \tilde{a}_1$ and $a_3 \leftrightarrow \tilde{a}_2$, and denote the arc length from point $a_2$ to $a_3$ by $s$, and that from point $\tilde{a}_1$ to $\tilde{a}_2$ is $\tilde{s}$, then Theorem 2 tells us that $\tilde{s} = \gamma s \Rightarrow \dfrac{s}{\tilde{s}} = \dfrac{1}{\gamma}$. Furthermore, with $\mathcal{C}$ being the boundary of $\mathcal{S}$; $s_t$ denoting the total arc length of $\mathcal{C}$; $\tilde{\mathcal{C}}$ representing the boundary of $\mathcal{S}$; and $s_t$ standing for the total arc length of $\mathcal{C}$, we also have $\tilde{s}_t = \gamma s_t \Rightarrow \dfrac{s_t}{\tilde{s}_t} = \dfrac{1}{\gamma}$. Thus, we find $ \dfrac{s}{\tilde{s}} = \dfrac{1}{\gamma} = \dfrac{s_t}{\tilde{s}_t}$. So, we see $\dfrac{s}{\tilde{s}} = \dfrac{s_t}{\tilde{s}_t}$ implying $\dfrac{s}{s_t} = \dfrac{\tilde{s}}{\tilde{s}_t}$. Consequently, we have $(\dfrac{s}{s_t} - \dfrac{\tilde{s}}{\tilde{s}_t})^2 = 0$. It is this final ``equality to zero" that forms the basis of our method. In particular, notice that since our choice of the second and third members of $\mathcal{F}$ was arbitrary, it follows that the ``equality to zero" holds for all consecutive members of $\mathcal{F}$. We formalize this observation below.
Consider any $a_i \in A$ (set $A$ is as defined above) and let us define the ordered set $A_{a_i} = \{ a_i, a^1,a^2, \dots,a^{L-1}\}$, such that $A_{a_i}$ captures the order in which points in set $A$ are to be encountered on the curve $\mathcal{C}$, when $\mathcal{C}$ is traversed in a counter clockwise manner starting out at $a_i$. With this, we can define an arc length set, $B_{a_i} = \{ s_{a_i}^1,s_{a_i}^2, \dots,s_{a_i}^L\}$, in which $s_{a_i}^k$, $k \in \{1,2, \dots, L-1 \}$ is the arc length between $a^k$ and its predecessor in the ordered set $A_{a_i}$. Notice that this leaves $s_{a_i}^L$ undefined. We simply define $s_{a_i}^L$ as the arc length from $a^L$ back to $a_i$. Analogously, for curve $\tilde{\mathcal{C}}$, we define $\tilde{A}_{a_j} = \{ \tilde{a}_j, \tilde{a}^1,\tilde{a}^2, \dots,\tilde{a}^{L-1}\}$ and $B_{a_j} = \{ \tilde{s}_{a_j}^1,\tilde{s}_{a_j}^2, \dots,\tilde{s}_{a_j}^L\}$. With these, we compute $C(a_i, \tilde{a}_j)$, the arc length correspondence score between point $a_i \in A \subset \mathcal{C}$ and point $\tilde{a}_j \in \tilde{A} \subset \tilde{\mathcal{C}}$ as follows:
\begin{equation} 
C(a_i, \tilde{a}_j) = \sum_{k=1} ^L (\dfrac { s_{a_i}^k }{{s}_t} - \dfrac{\tilde{s}_{\tilde{a}_j} ^ k}{\tilde{s}_t})^2
\end{equation}
From our earlier discussion, it should be clear that if the ordered sets $A_{a_i}$ and $\tilde{A}_{\tilde{a}_j}$ form a bijection, then $C(a_i, \tilde{a}_j)$ must equal zero. In this event, we can simply take points $a_i$ and $\tilde{a}_j$ as seed points for curves $\mathcal{C}$ $\tilde{\mathcal{C}}$ respectively. Having done so, the rest of the processing, towards obtaining scale and rotation invariant sample points, then becomes identical to Algorithm 1. With this, it is clear that we have been able to achieve a solution for the case wherein the functions $D(s)$ and $\tilde{D}(\tilde{s})$ each have more than one point of absolute minimum/maximum. Our solution is delineated in Algorithm 2 below. Again, while Algorithm 2 is able to efface the $n = 2^k$ requirement of the method presented in Section 4.1, it will still need some adaptations to handle digital curves. 
\subsection{Scale and Rotation Invariant Sampling By Critical Points of Shape Curvature }
In this section, we describe a technique for invariant shape boundary sampling using the local maxima of the curvature function of the shape. Indeed, Xie \textit{et. al.} (2007) have already described this sort of sampling technique, and they mentioned that it generally does not lead to equal-spaced sampling, unlike the distance minimization technique described in the previous section. However, they did not mention whether it is scale and rotation invariant or not. We will now prove that it indeed is scale and rotation invariant. Towards this, we prove two theorems. The first theorem captures the effect of scaling and rotation on curvature, while the second theorem underpins the effect of scaling and rotation on the critical points of functions, such as the curvature function, defined along shape boundaries: 
\newtheorem{thm5}[thm1]{Theorem} 
\begin{thm5}
Suppose $\mathcal{S}$ and $\tilde{\mathcal{S}}$ are two shapes in $\mathbb{R}^2$, such that $\tilde{\mathcal{S}} = R_{\phi} (\: \mathcal{E}_{\gamma}(\mathcal{S}) \: )$; and suppose $\mathcal{C}$ and $\tilde{\mathcal{C}}$ are twice differentiable simple closed curves delineating the boundaries of $\mathcal{S}$ and $\tilde{\mathcal{S}}$ respectively. Further, let $s$ be the arc length of $\mathcal{C}$, measured counter-clockwise from some point $p_i$ on $\mathcal{C}$; and let $\tilde{s}$ be the arc length of $\tilde{\mathcal{C}}$ measured counter-clockwise from the point $\tilde{p}_i = R_{\phi} (\: \mathcal{E}_{\gamma}(p_i) \:)$. Then, for any point $p_j$ on $\mathcal{C}$, the curvature vector at the point $R_{\phi} (\: \mathcal{E}_{\gamma}(p_j) \: )$ is $\gamma$ divided by the curvature vector at the point $p_i$. \\ \\
\end{thm5}
\begin{proof}[\textbf{\textit{Proof}}]
We set $\tilde{p}_i = \mathcal{S}_\gamma(p_i)$ and $\tilde{p}_j = \mathcal{S}_\gamma(p_j)$. The arc length at $p_j$ is denoted $s$ and that at $R_{\phi} (\: \mathcal{E}_{\gamma}(p_j) \: )$ is denoted $\tilde{s}$.
So, from Theorem 2, we know $s = \dfrac{1}{\gamma}\tilde{s}$. Differentiating this with respect to $\tilde{s}$, we get $s_{\tilde{s}} = \dfrac{1}{\gamma}$. Now one may write, $\tilde{\mathcal{C}}(\tilde{s}) = \tilde{\mathcal{C}}(s) = R_{\phi} ( \: \gamma x(s), \; \gamma y(s) \: )^T$. Differentiating with respect to $\tilde{s}$, we get: $\tilde{\mathcal{C}}_{\tilde{s}}(\tilde{s}) = \tilde{\mathcal{C}}_{\tilde{s}}(s) = R_{\phi}( \: \gamma x_s(s)s_{\tilde{s}}, \; \gamma y_s(s)s_{\tilde{s}} \: )^T = R_{\phi} ( \: x_s(s), \; y_s(s) \: )^T$, the last equality following from the fact that $s_{\tilde{s}} = \dfrac{1}{\gamma}$. Differentiating the above result with respect to $\tilde{s}$, we now find: $\tilde{\mathcal{C}}_{\tilde{s} \tilde{s}}(\tilde{s}) = \tilde{\mathcal{C}}_{\tilde{s} \tilde{s}}(s) = R_{\phi}( \: x_{ss}(s)s_{\tilde{s}}, \; y_{ss}(s)s_{\tilde{s}} \: )^T = R_{\phi}( \: \dfrac{1}{\gamma} x_{ss}(s), \; \dfrac{1}{\gamma}y_{ss}(s) \: )^T = R_{\phi} \dfrac{1}{\gamma}( \: x_{ss}(s), \; y_{ss}(s) \: )^T$. This proves the theorem, since $R_{\phi}( \: x_{ss}(s), \; y_{ss}(s) \: )^T$ is the curvature vector of $\mathcal{C}(s)$. In closing, we point out that the given proof makes implicit use of the fact that both $|\mathcal{C}_s(s)|$ and $|\tilde{\mathcal{C}}_{\tilde{s}}(\tilde{s})|$ are equal to unity.
\end{proof}
The second theorem now follows: 
\newtheorem{thm3}[thm1]{Theorem} 
\begin{thm3}
In $\mathbb{R}^2$, suppose $\mathcal{C}$ and $\tilde{\mathcal{C}}$ are twice differentiable simple closed curves, such that $\tilde{\mathcal{C}} = R_{\phi} (\: \mathcal{E}_{\gamma}(\mathcal{C}) \: )$; and suppose point $p_i \in \mathcal{C}$ and point $\tilde{p}_i \in \tilde{\mathcal{C}}$ are related by $\tilde{p}_i = R_{\phi} (\: \mathcal{E}_{\gamma}(p_i) \: )$. Further, for any two points $p_j \in \mathcal{C}$ and $\tilde{p}_j \in \tilde{\mathcal{C}}$ related by $\tilde{p}_j = R_{\phi} (\: \mathcal{E}_{\gamma}(p_j) \: )$, let $s$ be the arc length of $\mathcal{C}$, measured (counter) clockwise from $p_i$, and let $\tilde{s}$ be the arc length of $\tilde{\mathcal{C}}$, measured (counter) clockwise from $\tilde{p}_i$. Then, where $s_t$ is the total arc length of $\mathcal{C}$, and where $\tilde{s}_t$ is the total arc length of $\tilde{\mathcal{C}}$, if $G: [0, s_t]/ \{0,s_t\} \rightarrow \mathbb{R} $ and $\tilde{G}: [0, \tilde{s}_t]/ \{0,\tilde{s}_t\} \rightarrow \mathbb{R} $ are two functions satisfying $\tilde{G}(\tilde{s})= cG(s)$ for some positive constant, $c$, the following must hold. \textup{1).} $s_1$ is a minimum point of $G(s)$ if and only if $R_{\phi} (\: \mathcal{E}_{\gamma}(s_1))$ is a local minimum point of $\tilde{G}(\tilde{s})$. 
\textup{2).} $s_1$ is a maximum point of $G(s)$ if and only if $R_{\phi} (\: \mathcal{E}_{\gamma}(s_1))$ is a local maximum point of $\tilde{G}(\tilde{s})$.
\textup{3).} $s_1$ is an inflection point of $G(s)$ if and only if $R_{\phi} (\: \mathcal{E}_{\gamma}(s_1))$ is an inflection point of $\tilde{G}(\tilde{s})$. \\
\end{thm3} 
Given below, the proof of the above theorem makes use of the fact that if $s_1$ is a critical point of $D(s)$, and $D^k(s_1)$ is the first non-zero derivative of $D(s)$ at $s_1$, then $s_1$ is a local minimum/maximum point of $D(s)$ if $k$ is even and $D^k(s_1)$ is positive/negative. If $k$ is odd, then $s_1$ is an inflection point. \\
\begin{proof}[\textbf{\textit{Proof}}] \mbox{} \\
Firstly, observe that Theorem 2 above tells us that $s$ and $\tilde{s}$ are related by $\tilde{s} = \gamma s \Rightarrow s = \dfrac{1}{\gamma} \tilde{s}$. Differentiating with respect to $\tilde{s}$, we have $s_{\tilde{s}} = \dfrac{1}{\gamma}$. Now, differentiating $\tilde{G}(\tilde{s}) = cG(s) $ with respect to $\tilde{s}$, we write: $\tilde{G}_{\tilde{s}}(\tilde{s}) = cG_{\tilde{s}}(s) = cG_{s}(s)s_{\tilde{s}} = \dfrac{1}{\gamma}cG_{s}(s)$. Summarily, we have found $\tilde{G}_{\tilde{s}}(\tilde{s}) = \dfrac{1}{\gamma}cG_{s}(s)$. Since $c$ and $\gamma$ are non-zero, it follows that $\tilde{G}_{\tilde{s}}(\tilde{s}) = 0 $ if and only if $G_s(s) = 0$. This means, if we consider two points, $\tilde{s}_1 \in \tilde{I}$ and $s_1 \in I$, such that $\tilde{s}_1 = \gamma s_1$, then $\tilde{s}_1$ is a critical point of $\tilde{G}(\tilde{s})$ if and only if $s_1$ is a critical point of $G(s)$. To proceed, we differentiate $\tilde{G}_{\tilde{s}}(\tilde{s}) = \dfrac{1}{\gamma}cG_{s}(s)$ with respect to $\tilde{s}$, $\tilde{G}_{\tilde{s} \tilde{s}}(\tilde{s}) = \dfrac{1}{\gamma}cG_{s \tilde{s}}(s) = \dfrac{1}{\gamma}cG_{ss}(s)s_{\tilde{s}} = \dfrac{1}{\gamma^2}cG_{ss}(s)$. So, we have found, $\tilde{G}_{\tilde{s} \tilde{s}}(\tilde{s}) = \dfrac{1}{\gamma^2}cG_{ss}(s)$. Indeed, if one continues to differentiate, one will find: $\tilde{G}^k(\tilde{s}) = \dfrac{1}{\gamma^k}cG^k(s)s_{\tilde{s}}$, where $\tilde{G}^k(\tilde{s})$ is the $k$-th derivative of $\tilde{G}(\tilde{s})$ with respect to $\tilde{s}$, and ${G}^k(s)$ is the $k$-th derivative of $G(s)$ with respect to $s$. Now, since $c$ and $\gamma$ are positive, it follows that $\tilde{G}^k(\tilde{s}_1) > 0 $ if and only if ${G}^k(s_1) > 0$. Similarly, $\tilde{G}^k(\tilde{s}_1) < 0 $ if and only if ${G}^k(s_1) < 0$. Also, $\tilde{G}^k(\tilde{s}_1) = 0 $ if and only if ${G}^k(s_1) = 0$. Thus, $\tilde{G}^k(\tilde{s_1})$ is the first positive/negative/zero derivative of $\tilde{G}(\tilde{s})$ at $\tilde{s}_1$ if and only if ${G}^k(s_1)$ is the first positive/negative/zero derivative of $G(s)$ at $s_1$. In particular, if $k$ is even and $\tilde{G}^k(\tilde{s_1})$ is the first positive derivative of $\tilde{G}(\tilde{s})$ at $\tilde{s_1}$, then $\tilde{s}_1$ is a local minimum point of $\tilde{G}(\tilde{s})$ and ${s}_1$ is a local minimum point of ${G}(\tilde{s})$. On the flip side, if $k$ is even and $\tilde{G}^k(\tilde{s_1})$ is the first negative derivative of $\tilde{G}(\tilde{s})$ at $\tilde{s_1}$, then $\tilde{s}_1$ is a local maximum point of $\tilde{G}(\tilde{s})$ and ${s}_1$ is a local maximum point of ${G}(\tilde{s})$. Finally, if $k$ is odd and $\tilde{G}^k(\tilde{s_1})$ is the first non-zero derivative of $\tilde{G}(\tilde{s})$ at $\tilde{s_1}$, then $\tilde{s}_1$ is an inflection point of $\tilde{G}(\tilde{s})$ and ${s}_1$ is an inflection point of ${G}(\tilde{s})$
\end{proof}
We can now explore ways of exploiting the above theorems for the task at hand. First, notice from Theorem 4, that if $\kappa(s)$ is the curvature function on $\mathcal{C}(s)$ and $\tilde{\kappa}(\tilde{s})$ is the curvature function on $\tilde{\mathcal{C}}(\tilde{s})$, then we must have $\tilde{\kappa}(\tilde{s})$ = $c \kappa(s)$ where $c = \dfrac{1}{\gamma}$. Thus, $\kappa(s)$ and $\tilde{\kappa}(\tilde{s})$ can respectively play the roles of $Z(s)$ and $\tilde{Z}(\tilde{s})$ in Theorem 5. So, according to Theorem 5, we see that $\kappa(s)$ and $\tilde{\kappa}(\tilde{s})$ must have the same number of points of local maximum. Let this number of points be $L$. If $L = 1$, we simply take this single point of local maximum as the seed point and proceed as described in Section 4.2. However, if $L > 1$, then we must solve an arc length correspondence problem embodied by Equation 3. This allows us to compute $L$ scale and rotation invariant sample points around $\mathcal{C}$ and $\tilde{\mathcal{C}}$. Now, suppose we need $n$ sample points, and we found that $n \neq L$. If $n < L$, we simply discard $L - n$ of the already computed points. But, if $n > L$, then we must compute more sample points. One way to do this is to look at the smallest interval on each curve, and then use the length of this interval to mark off more points on the curve, beginning from the largest interval to the smallest one, until a total of $n$ sample points can be found on the curve. We describe an algorithm that encompasses our local curvature maxima sampling technique as Algorithm 3 below.S 
\section{Theories and Methods for Achieving Invariance with Compact Shape Trees: Spatial Domain }
In the first place, translation invariance is an innate characteristic of compact shape trees. Moreover, in both the spatial and frequency domains, compact Shape trees lend themselves to simple techniques for achieving scale and rotation invariance. Beginning with the spatial domain, we detail these techniques in this section. Let $\mathcal{T}_{p_i} = \langle \mathcal{V}_{p_i}, f_1,f_2,...,f_m \rangle$ be a compact shape tree. We consider first how to achieve rotation invariance for the vector tree, $\mathcal{V}_{p_i}$, noting that rotation does not affect the moduli of the vectors in $\mathcal{V}_{p_i}$, but definitely does affects their angles. In vector notation, suppose $\mathcal{V}_{p_i} = \{v_1,v_2, ...v_{n-1}\}$. To achieve rotation invariance, we simply take the differences of the angles of consecutive vectors in $\mathcal{V}_{p_i}$. Formally, assuming each $v_i$ is in polar form, making angle $\theta_i$ with the positive $x$-axis, and having magnitude $r_i$, it should be obvious that the ordered set $S_\alpha = \{\alpha_{i+1} - \; \alpha_{i} \mid \; i \in \mathbb{Z}^+, i \leq n-1, \alpha_n = \alpha_1 \}$ is rotation invariant. For instance, when the underlying shape is rotated through angle $\phi$, each vector, $v_i$, in $\mathcal{V}_{p_i}$ also rotates through $\phi$. If one denotes the new angle of $v_i$, after the rotation, by $\tilde{\alpha}_i = \alpha_i + \phi$, and defines the set $S_{\tilde{\alpha}} = \{\tilde{\alpha}_{i+1} - \; \tilde{\alpha}_{i} \mid \; i \in \mathbb{Z}^+, i \leq n-1, \alpha_n = \alpha_1 \}$, one easily finds $S_{\tilde{\alpha}} = S_{\alpha}$.
It remains to discuss rotation invariance for the features $f_1, f_2, ..., f_m$ in $\mathcal{T}_{p_i}$. From a previous discussion, we have seen that these features can either be vectors or scalars. Analysis for the case when an $f_i$ is a vector is similar to the analysis for vector trees above. We deal only with the case wherein rotation preserves the angle made by vector $f_i$ and the underlying shape boundary's tangent direction at the sample point, $p_i$, where $f_i$ is measured. It turns out this is the case for the vectorial features, such as the curvature vector and the tangent vector, we are interested in; the curvature vector will always make an angle of $\pm 90 $ degrees with the tangent vector, while the tangent vector will of course always make an angle of zero degrees with itself. Under this restriction, it can be shown that when the underlying shape rotates, say, through $\phi$ the vectorial feature $f_i$ also rotates through $\phi$. Thus, we can again achieve rotation invariance for each vectorial $f_i$ by taking the consecutive differences of its angles, as was done for vector trees above. The case when a feature $f$ is particularly easy to handle. We first bring to mind the point notation $\mathcal{V}_{p_i} = \{p_i, p^1, p^2, ..., p^{n-1}\}$. With this, one sees that, where $f_{p_i}$ is the value of feature $f$ at point $p_i$, and $f^i$ is the value of $f$ at $p^i$, then the ordered set $S_f = \{f_{p_i}, f^1, f^2, ..., f^{n-1}\}$ is rotation invariant.
We now turn to the topic of scale invariance in the spatial domain. Clearly, scaling does not affect the angles of the vectors in $\mathcal{V}_{p_i} = \{v_1,v_2, ...v_{n-1}\}$, but does affects their moduli by scaling them. A well known strategy for achieving scale invariance in this kind of situation is to divide each member in the collection of these moduli by the maximum of the collection (Alajlan, Wang). But, we must ask about the features in $\mathcal{T}_{p_i}$. Although we restrict attention to only the curvature vector (and its magnitude), we still need a lemma to proceed:
\section{Theories and Methods for Achieving Invariance with Compact Shape Trees: Frequency Domain}
As mentioned in our introduction, compact shape trees lend themselves to analysis in the frequency domain. In particular, it is possible to extract a scale and rotation invariant version of the underlying vector tree, $\mathcal{V}_{p_i}$, in any compact shape tree. In what follows, we explicitly separate the ordered set $\mathcal{V}_{p_i} = \{v_1,v_2, \dots, v_{n-1}\}$ into two ordered sets, $\varPhi = \{e^{\theta_1}, e^{\theta_2}, \dots, e^{\theta_{n -1}} \}$, and $\varGamma = \{l_1, l_2, \dots, l_{n-1} \}$, where $\theta_i$ is the angle the vector $v_i \in \mathcal{V}_{p_i}$ makes with the positive x-axis, and $l_i$ is the length of $v_i$. We may consider the aperiodic discrete Fourier transform, $X_{\theta}(\omega)$, of the sequence of exponentiated angles in $\varPhi$, and write:
\begin{equation}
X_{\theta}(\omega) = \sum_{i = 1} ^ {n-1} e^ {\theta_i} e ^{-j \omega i}
\end{equation}
where $j = \sqrt{-1}$
Analogously, we may denote the aperiodic discrete Fourier transform of the sequence of moduli in set $\varGamma$ by $X_{l}(\omega)$, and write: 
\begin{equation}
X_{l}(\omega) = \sum_{i = 1} ^ {n-1} l_i \, e ^{-j \omega i}
\end{equation}
At this juncture, bring to mind the physical fact that rotation affects only the angles in a vector tree; as far as rotation is concerned, we need only bother about the ordered set $\varPhi = \{e^{\theta_1}, e^{\theta_2}, \dots, e^{\theta_{n -1}} \}$. Using Equation 11, we can state and prove a theorem about the rotation invariance of an arbitrary vector tree, $\mathcal{V}_{p_i}$:
\newtheorem{thm6}[thm1]{Theorem} 
\begin{thm6}
Suppose $\mathcal{S}$ and $\tilde{\mathcal{S}}$ are two shapes such that $\tilde{\mathcal{S}}$ is the result of rotating $\mathcal{S}$ through angle $\phi$ in the plane; and let $X_{\theta}(\omega)$ and $\tilde{X}_{\theta}(\omega)$ be the aperiodic discrete Fourier transforms associated with $\tilde{\mathcal{S}}$ and $\tilde{\mathcal{S}}$ respectively. Then, for any $\omega$, we have $\tilde{X}_{\theta}(\omega) = e^{\phi}X_{\theta}(\omega)$ and for any $\omega_1, \omega_2$, we have $\dfrac{{X}_{\theta}(\omega_1)}{{X}_{\theta}(\omega_2)} = \dfrac{\tilde{X}_{\theta}(\omega_1)}{\tilde{X}_{\theta}(\omega_2)}$, which says that the quantity $\dfrac{{X}_{\theta}(\omega_1)}{{X}_{\theta}(\omega_2)}$ is a rotation invariant property of the shape $\mathcal{S}$. 
\end{thm6}
\begin{proof} [\textbf{Proof}] \mbox{} \\
The sequence of exponentiated angles associated with the vector tree of shape $\mathcal{S}$ is in the ordered set $\varPhi = \{e^{\theta_1}, e^{\theta_2}, \dots, e^{\theta_{n -1}} \}$. When $\mathcal{S}$ is rotated through $\phi$ degrees, each $\theta_i$ must increase by $\phi$. Hence, the ordered set of exponentiated angles associated with shape $\tilde{\mathcal{S}}$ is $\tilde{\varPhi} = \{e^{\phi + \theta_1 }, e^{\phi + \theta_2}, \dots, e^{\phi + \theta_{n -1} } \}$. Now, the Fourier transform of this sequence is $\tilde{X}_{\theta}(\omega) = \sum_{i = 1} ^ {n-1} e^ {\phi + \theta_i} e ^{-j \omega i} = \sum_{i = 1} ^ {n-1} e^ {\phi} e^{ \theta_i} e ^{-j \omega i} = e^ {\phi} \sum_{i = 1} ^ {n-1} e^{ \theta_i} e ^{-j \omega i} = e^ {\phi} X_{\theta}(\omega)$. From this, we see that $\tilde{X}_{\theta}(\omega) = e^ {\phi} X_{\theta}(\omega)$, for any $\omega$. Thus, for a specific $\omega_1$, we may write $\tilde{X}_{\theta}(\omega_1) = e^ {\phi} X_{\theta}(\omega_1) \: \Rightarrow \: \dfrac{\tilde{X}_{\theta}(\omega_1)}{X_{\theta}(\omega_1)} = e^ {\phi}$. Similarly, for another specific $\omega_2$, one finds $\dfrac{\tilde{X}_{\theta}(\omega_2)}{X_{\theta}(\omega_2)} = e^ {\phi}$. Consequently, one sees, $e^ {\phi} = \dfrac{\tilde{X}_{\theta}(\omega_2)}{X_{\theta}(\omega_2)} = \dfrac{\tilde{X}_{\theta}(\omega_1)}{X_{\theta}(\omega_1)} \: \Rightarrow \: \dfrac{{X}_{\theta}(\omega_1)}{{X}_{\theta}(\omega_2)} = \dfrac{\tilde{X}_{\theta}(\omega_1)}{\tilde{X}_{\theta}(\omega_2)}$. 
\end{proof}
\noindent \newtheorem{thm7}{Remark} 
\begin{thm7}
The above proof should reveal why we exponentiated the angles before taking their Fourier transform. This is a deliberate arrangement aimed at achieving rotation invariance in the manner described by Theorem 6. Fortunately, the function $f(\theta) = e^\theta$ turns out to be a good choice for this purpose, because it is a one-to-one function. To see why we need a one-to-one function here, consider $g(\theta) = \sin\theta$ which is not one-to-one, and notice that it diminishes the discriminatory power of the sequence, $\theta_1, \theta_2, \dots, \theta_n$, as a feature set; for instance, $g(\theta)=\sin\theta$ maps both $\frac{\pi}{4}$ and $\frac{3\pi}{4}$ to the same value! 
\end{thm7}
We turn to the issue of scale invariance. Scaling affects only the lengths of the vectors in a vector tree. So, as far as scaling is concerned, we need only worry about the ordered set, $\varGamma = \{l_1, l_2, \dots, l_{n-1} \}$. The following theorem tells us how to achieve scale invariance for a vector tree: 
\newtheorem{thm8}[thm1]{Theorem} 
\begin{thm8}
Suppose $\mathcal{S}$ and $\tilde{\mathcal{S}}$ are two shapes such that $\tilde{\mathcal{S}}$ is the result of scaling $\mathcal{S}$ by a factor of $\gamma$ ; and let $X_{l}(\omega)$ and $\tilde{X}_{l}(\omega)$ be the aperiodic discrete Fourier transforms associated with $\tilde{\mathcal{S}}$ and $\tilde{\mathcal{S}}$ respectively. Then, for any $\omega$, we have $\tilde{X}_{l}(\omega) = \gamma X_{l}(\omega)$ and for any $\omega_1, \omega_2$, we have $\dfrac{{X}_{l}(\omega_1)}{{X}_{l}(\omega_2)} = \dfrac{\tilde{X}_{l}(\omega_1)}{\tilde{X}_{l}(\omega_2)}$, which says that the quantity $\dfrac{{X}_{l}(\omega_1)}{{X}_{l}(\omega_2)}$ is a scale invariant property of the shape $\mathcal{S}$. 
\end{thm8}
\begin{proof} [\textbf{Proof}] \mbox{} \\
The sequence of lengths associated with the vector tree of shape $\mathcal{S}$ is in the ordered set $\varGamma = \{l_1, l_2, \dots, l_{n -1} \}$. When $\mathcal{S}$ is scaled by a factor of $\gamma$, each $l_i$ must be scaled by $\gamma$. Hence, the ordered set of lengths associated with shape $\tilde{\mathcal{S}}$ is $\tilde{\varGamma} = \{\gamma l_1, \gamma l_2, \dots, \gamma l_{n -1} \}$. Now, the Fourier transform of this sequence is $\tilde{X}_{l}(\omega) = \sum_{i = 1} ^ {n-1} \gamma l_i e ^{-j \omega i} = \gamma \sum_{i = 1} ^ {n-1} l_i e ^{-j \omega i} = \gamma X_{l}(\omega)$. From this, we see that $\tilde{X}_{l}(\omega) = \gamma X_{l}(\omega)$, for any $\omega$. Thus, for a specific $\omega_1$, we may write $\tilde{X}_{l}(\omega_1) = \gamma X_{l}(\omega_1) \: \Rightarrow \: \dfrac{\tilde{X}_{l}(\omega_1)}{X_{l}(\omega_1)} = \gamma $. Similarly, for another specific $\omega_2$, one finds $\dfrac{\tilde{X}_{l}(\omega_2)}{X_{l}(\omega_2)} = \gamma$. Consequently, one sees, $\gamma = \dfrac{\tilde{X}_{l}(\omega_2)}{X_{l}(\omega_2)} = \dfrac{\tilde{X}_{l}(\omega_1)}{X_{l}(\omega_1)} \: \Rightarrow \: \dfrac{{X}_{l}(\omega_1)}{{X}_{l}(\omega_2)} = \dfrac{\tilde{X}_{l}(\omega_1)}{\tilde{X}_{l}(\omega_2)}$. 
\end{proof}
\end{document}